\documentclass[12pt]{article}
\usepackage{amsthm,amsmath,amssymb,stackengine,comment}
\usepackage{amsfonts,mathrsfs}
\usepackage{color}
\usepackage{enumerate}
\usepackage{booktabs,float}
\usepackage[flushleft]{threeparttable}
\usepackage[figuresright]{rotating}
\usepackage{graphicx}
\usepackage{multirow}
\usepackage{diagbox}
\usepackage{setspace}
\usepackage{enumerate}
\usepackage{mathtools}
\usepackage{subfigure}
\usepackage{algorithm}
\usepackage{algorithmicx}
\usepackage[noend]{algpseudocode}
\usepackage{natbib}
\usepackage{url} 
\usepackage{xr}

\newcommand{\blind}{1}

\theoremstyle{plain}

\newtheorem{assumption}{Assumption}

\newtheorem{theorem}{Theorem}[section]
\newtheorem{corollary}{Corollary}[section]

\newtheorem{Remark}{Remark}
\theoremstyle{remark}
\newtheorem{definition}{Definition}

\newcommand{\btheta}{\boldsymbol{\theta}}
\newcommand{\balpha}{\boldsymbol{\alpha}}

\newcommand{\bmu}{\boldsymbol{\mu}}

\newcommand{\bsigma}{\boldsymbol{\sigma}}
\newcommand{\argmin}{\mathop{\mathrm{arg\,min}}}
\date{}

\begin{document}

	\def\spacingset#1{\renewcommand{\baselinestretch}%
		{#1}\small\normalsize} \spacingset{1}

	
	\if1\blind
	{
		\title{\bf A Conditional Distribution Equality Testing Framework using Deep Generative Learning}
		\author{
			{  Siming Zheng\textsuperscript{1}, Tong Wang\textsuperscript{1},
				Meifang Lan\textsuperscript{2}, and Yuanyuan Lin\textsuperscript{2}}\\ 
			\small{\textsuperscript{1}Department of Industrial Systems Engineering and Management, }\\ \small{Nationalal University of Singapore, Singapore}\\
			\small{\textsuperscript{2}Department of Statistics and Data Science, } \\ \small{The Chinese University of Hong Kong, Hong Kong, China}\\
		}
		
		\footnotetext[2]{Yuanyuan Lin is the corresponding author (ylin@sta.cuhk.edu.hk). The first three authors are co-first authors.
		}
		\maketitle
	} \fi
	
	\if0\blind
	{
		\bigskip
		\bigskip
		\bigskip
		\begin{center}
			{\LARGE\bf A Conditional Distribution Equality Testing Framework using Deep Generative Learning}
		\end{center}
		\medskip
	} \fi

	\bigskip
	\begin{abstract}
		In this paper, we propose a general framework for  testing the conditional distribution equality in a two-sample problem, which is most relevant to covariate shift and causal discovery.  
		Our framework is built on neural network-based generative methods and sample splitting techniques by transforming the conditional testing problem into an unconditional one. 
		We introduce the generative classification accuracy-based conditional distribution equality test (GCA-CDET) to illustrate the proposed framework. We  establish
		the convergence rate for the learned generator by   
		deriving new results related to the recently-developed offset Rademacher complexity and prove the testing consistency of GCA-CDET under mild conditions.
		Empirically, we  conduct numerical studies including  synthetic datasets and two real-world datasets, demonstrating the effectiveness of our approach. Additional discussions on the optimality of the proposed framework are provided in the online supplementary material.
	\end{abstract}

	\noindent%
	{\it Keywords:}  Conditional Distribution Equality Testing; Deep Neural Networks; Generative Learning; Two-sample Testing.
	\vfill
	
	\newpage
	\spacingset{1.9} 
	
	\section{Introduction}
	
	In this paper, we study a basic  statistical testing problem: determining whether the conditional distributions of two datasets are the same. Given two independent random samples $\mathbb{D}_1=\{(Y_{1,i},X_{1,i})\}_{i=1}^{n_1}$ and $\mathbb{D}_2=\{(Y_{2,i},X_{2,i})\}_{i=1}^{n_2}$, our central concern is to test
	\begin{equation}\label{cdt_problem}
		H_0: \mathbb{P}_{1,Y|X}=\mathbb{P}_{2,Y|X} \quad \text { v.s } \quad H_1: \mathbb{P}_{1,Y|X}\neq\mathbb{P}_{2,Y|X},
	\end{equation}
	where $\mathbb{P}_{k,Y|X}$ is the conditional distribution of $Y$ given $X$ for the dataset $\mathbb{D}_k,k=1,2$. This problem has been receiving increasing attention in recent years due to the popularity of covariate shift \citep{Quionero2009,pathak2022new,qiu2024efficient} and causal inference \citep{peters2016causal,lei2021conformal,fan2024environment} studies. 
	
	In the context of covariate shift, source and target datasets are available, each containing response variables and covariates.
	The covariate shift setting  assumes that the conditional distributions of the response given the covariates remain invariant across datasets, but the marginal covariate distributions may vary.  For example, \cite{qiu2023prediction} examined HIV risk prediction using a South African cohort dataset from \cite{tanser2013high} and documented that the empirical distribution of HIV prevalence varies between the source populations (urban and rural communities) and the target population (peri-urban communities), whereas the conditional distributions of the outcome given covariates \textit{seem similar} across these populations, which consequently suits to study the prediction sets adaptive to unknown covariate shift. Hence, the conditional equality test may provide preliminary guideline for such an analysis.
	
	Conditional equality testing plays an important role in causal discovery, especially in invariance detection. It is now well-established that  conditional distribution invariance is helpful for detecting causal variables \citep{peters2016causal,buhlmann2020invariance,kook2025model,varici2025score}. Specifically, a covariate $X$ constitutes a causal parent of $Y$ if the conditional distributions of $Y$ given $X$ remain stable across different environments. Consequently, testing the equality of conditional distributions across environments may provide insights into the causal relationships involving $X$.
	
	To our knowledge, while the testing problem (\ref{cdt_problem}) is fundamental, existing literature directly addressing it remains limited. Drawing on the connection established in Remark 4 of \cite{Shah:Peters:hardness2020} between the testing problem (\ref{cdt_problem}) and conditional independence testing, a straightforward approach involves constructing a pseudo-cluster-augmented dataset $\widetilde{\mathbb{D}}_k=\{(Y_{k,i},X_{k,i},W_{k,i})\}_{i=1}^{n_k}$, where $W_{k,i}=k$ for $k=1,2$, and applying conditional independence tests to the merged data $\widetilde{\mathbb{D}}_1\cup \widetilde{\mathbb{D}}_2$ to assess whether $Y\perp W|X$ holds. Under this formulation, conditional independence implies that the conditional distributions $Y|X,W=1$ and $Y|X,W=2$ are equivalent. Existing conditional independence tests include: (1) kernel-based methods \citep{zhang2011kernel,wang2015conditional}, which may be ineffective in high-dimensional covariate settings; (2) permutation-based approaches \citep{berrett2020conditional}; and (3) generative-learning-based methods \citep{bellot2019conditional,shi2021double,zhang2025doubly}. Although these tests offer valuable insights, their lack of specificity to the target problem (\ref{cdt_problem}) may compromise performance in highly imbalanced scenarios, e.g. when $n_2\ll n_1$. For generative-learning-based conditional independence tests, in particular, the influence of the smaller dataset may be substantially diminished, as distribution learning relies on merged data where observations from the smaller dataset are relatively sparse compared to the larger one. Meanwhile, imbalanced scenarios are not rare in practice: In many contemporary  transfer learning applications under covariate shifts \citep{shimodaira2000improving,pathak2022new}, the source data are often abundant while the target data are limited, a primary motivation for leveraging source data information.  Lately,  a recently-accepted paper \citep{Tian12112025}  also highlighted the importance and challenges of imbalanced scenarios in data synthesis augmentation, and developed a generative augmentation framework using deep generative learning to address this problem.
	
	Recently,  \cite{hu2023two} and \cite{chen2024biased} proposed two novel and conceptually appealing tests on the testing problem (\ref{cdt_problem}) based on the estimation of  the conditional-density ratio as well as the marginal-density ratio. These ratio estimation could be challenging and unstable, especially when the dimension of the covariates is high. Moreover, as shown by \cite{kanamori09a,Kanamori2012StatisticalAO,kato2021nonnegative}, the convergence rates for density-ratio estimates usually depend on the smaller sample size of the two samples, which again sheds light on that  these methods may not be effective  in imbalanced cases.
	
	To address the challenge in imbalanced cases, in this paper, we leverage tools from modern generative learning and the data splitting technique to develop a general and flexible framework that can, in principle, incorporate any  conditional generative learning method and two-sample testing approach. State-of-the-art conditional generative learning methods include  conditional generative adversarial networks \citep{zhou2023deep}, mixture density networks \citep[MDNs,][]{zhou2023testing}, conditional flow matching, and conditional diffusion model \citep{lipman2023flow}, among many others. Regarding two-sample testing approaches, they include: (1) rank-based tests \citep{vardi2000multivariate,Rousson2002},  integral probability metric-based tests \citep{baringhaus2004new,szekely2004testing,Ramdas:Trillos:Cuturi:2017} and so on.  
	Unlike the aforementioned methods, our proposed method can avoid the delicate density-ratio estimation or distribution learning using the merged data, thus it 
	can still work well  for imbalanced data.
	\subsection{Our contributions} 
	
	In this paper we develop a  general and flexible 
	framework to  test  the equality of two conditional distributions. In its core, we learn the conditional generator based on the larger dataset and employ the well-learned generator to the other dataset.  Through data splitting, we would transform the challenging conditional two sample testing question into the well-studied unconditional two-sample testing problem. Our framework can accommodate both multivariate response and high-dimensional covariates, and it is particularly useful when the two samples are  imbalanced. In principle, our framework can incorporate any state-of-the-art conditional generative learning method and two-sample testing approach. As an illustration, we  utilize MDNs similar to \cite{zhou2023testing}  to estimate 
	the conditional density, based on which a conditional generator can be learned, and then propose a generative classification accuracy-based conditional distribution equality test (GCA-CDET), which is particularly computationally efficient. Under mild conditions, we proved the testing consistency of GCA-CDET. Finally, we compute  GCA-CDET in numerical experiments to examine its finite-sample performance, which provides supporting evidence for the effectiveness of our method, especially in imbalanced cases.
	
	Regarding the optimality of the proposed framework, in the supplementary material, we have established a minimax lower bound for  testing  the equality of two conditional distributions under certain smoothness conditions and demonstrated that a generative permutation-based conditional distribution equality test can attain the minimax lower bound (Theorem S.4 \& Corollary S.5), and its adaptive version achieves the minimax lower bound up to an iterated logarithm factor (Theorem S.6).
	
	\subsection{Paper organization and notation} The rest of the paper is organized as follows.  In Section \ref{CD-Testing}, we first describe the  problem setup and the motivation for the proposed testing framework. We then introduce our proposed testing procedure. Section \ref{specific_test} introduce the mixture density networks
	to learn the conditional generator and the generative classification accuracy-based conditional distribution equality test (GCA-CDET).  In Section \ref{theory}, we 	
	derive the convergence rate for the estimated conditional generator and prove the testing consistency of GCA-CDET under proper conditions. In Section \ref{sec:simu}, we conduct  simulation studies, and in Section \ref{sec:data}, we conduct real data analysis on two datasets. A few concluding remarks and discussions  are given in Section \ref{Discussion}.
	
	Throughout the paper,  for two positive deterministic sequences $a_n$ and $b_n$, we denote $a_n \asymp b_n$ if $c<a_n / b_n<C$ for some absolute constants $c, C>0$ and all $n$ larger than some $n_0$. The symbol $\|\cdot\|_i$ refers to the $L_i$ norm w.r.t. the Lebesgue measure and $\mathbb{I}[\cdot]$ denotes the standard 0-1 indicator function.  For any $N\in\mathbb{N}^+$, we use $[N]$ to denote the set $\{1,2,\ldots,N\}$,  $\lceil a \rceil$ and $\lfloor a\rfloor$ to denote the smallest integer no less than $a$ and the largest integer smaller than $a$, respectively, where $a\in\mathbb{R}$.
	
	\section{Testing for the equality of two conditional distributions}\label{CD-Testing}
	Let $X\in\mathcal{X}\subseteq\mathbb{R}^{d}$ and $Y\in\mathcal{Y}\subseteq\mathbb{R}^{p}$ be the multivariate covariate and  response random vectors, respectively, where $\mathcal{X},\mathcal{Y}$ are the corresponding measurable spaces, and $d,p\in\mathbb{N}^+$ are the respective dimensions. Recall that there are two independent random samples $\mathbb{D}_1=\{(Y_{1,i},X_{1,i})\}_{i=1}^{n_1}$ and $\mathbb{D}_2=\{(Y_{2,i},X_{2,i})\}_{i=1}^{n_2}$, that are independent and identically distributed (i.i.d.) observations from the unspecified joint distributions $\mathbb{P}_{1,Y,X}$ and $\mathbb{P}_{2,Y,X}$ on $\mathcal{Y}\times\mathcal{X}$, respectively. We use $\mathbb{P}_{1,Y|X}$ to denote the conditional distribution of $Y$ given $X$ under $\mathbb{P}_{1,Y,X}$, and let $\mathbb{P}_{1, X}$ be the corresponding marginal distribution of $X$. 
	Accordingly,  we use $\mathbb{P}_{1,Y|X=x}$ denote the conditional distribution of $Y$ given $X=x$.
	And  $\mathbb{P}_{2,Y|X},\mathbb{P}_{2, X}$ and $\mathbb{P}_{2,Y|X=x}$ are defined analogously.
	
	Given the data $\mathbb{D}_1$ and $\mathbb{D}_2$,  our goal is to test whether the two conditional distributions $\mathbb{P}_{1,Y|X}$ and $\mathbb{P}_{2,Y|X}$ are identical based on the two samples $\mathbb{D}_1$ and $\mathbb{D}_2$, the testing problem (\ref{cdt_problem}) presented in the introduction. Given that the testing problem (\ref{cdt_problem}) presents substantial challenges  when the dimension of $X$ is moderate to high, and all related distributions including the unconditional and conditional ones are continuous, 
	our discussion below is concerned with such challenging scenarios.
	\subsection{Motivation}\label{Motivation}
	Suppose that  the marginal distributions of $X$ are the same across the two samples, i.e. $\mathbb{P}_{1, X}=\mathbb{P}_{2, X}$, the testing problem in (\ref{cdt_problem}) reduces to an unconditional two-sample testing problem:
	\[
	\tilde{H}_0: \mathbb{P}_{1,Y,X}=\mathbb{P}_{2,Y,X}\quad \text { v.s } \quad\tilde{H}_1:  \mathbb{P}_{1,Y,X}\neq\mathbb{P}_{2,Y,X}.
	\]
	Then, those existing two-sample testing methods mentioned in the introduction can be used to test the equality of the two conditional distributions.
	However,  the ideal assumption that $\mathbb{P}_{1, X}=\mathbb{P}_{2, X}$ is often violated  in many practical situations, 
	e.g. in the presence of covariate shift.
	\begin{Remark}
		Generally, when $\mathbb{P}_{1, X}\neq\mathbb{P}_{2, X}$ and both distributions are continuous, the classic unconditional two-sample tests are not directly applicable to the conditional equality testing problem (\ref{cdt_problem}). To see this, suppose for a given $x\in\mathcal{X}$, when the conditional distributions $\mathbb{P}_{1,Y|X=x}$ and $\mathbb{P}_{2,Y|X=x}$ are continuous, the data points with $X=x$ in the available datasets $\mathbb{D}_1$ and $\mathbb{D}_2$ are extremely sparse; in other words, the probability of observing data  with $X=x$ simultaneously in both datasets  approaches zero. Consequently, given $X=x$,
		there are insufficient observations available for the response variable to apply the classical unconditional two-sample tests for assessing whether $\mathbb{P}_{1,Y|X=x}=\mathbb{P}_{2,Y|X=x}$.
	\end{Remark}
	
	In the following, we show how to  turn the conditional distribution testing problem in (\ref{cdt_problem})  into an unconditional one,  with the help of  neural network-based generative methods and sample splitting techniques.  To motivate our proposed framework, suppose we have prior knowledge on how to draw samples $Y$ from the conditional distribution $\mathbb{P}_{1,Y|X}$, that is we have complete information about  $\mathbb{P}_{1,Y|X}$.  Without loss of generality, let $\eta$ follow a continuous distribution that is easy to sample from, e.g. the uniform distribution $\text{Unif}[0,1]$. And suppose that there is a known  function $V$ satisfying that for fixed $x$,
	\begin{equation}
		\label{gen}
		V(x,\eta)\sim \mathbb{P}_{1,Y|X=x}. 
	\end{equation}
	In fact,  the existence of such a generator function $V$ is guaranteed by the noise outsourcing lemma \citep[Lemma 3.1,][]{austin2015exchangeable} under mild conditions. 
	Without loss of generality, assume that $n_2$ is an even number. Then, we  randomly partition the second sample  $\mathbb{D}_2=\{(Y_{2,i},X_{2,i})\}_{i=1}^{n_2}$ into two equal-size sub-datasets
	\[
	\mathbb{D}_{21}=\{(Y_{21,i},X_{21,i})\}_{i=1}^{n_2/2}\quad\text{and}\quad \mathbb{D}_{22}=\{(Y_{22,i},X_{22,i})\}_{i=1}^{n_2/2}.
	\]
	Let the random noises $\eta_{1},\eta_{2},\ldots,\eta_{n_2/2}\stackrel{\text { i.i.d }}{\sim} \text{Unif}[0,1]$ be generated independently from $\mathbb{D}_1$ and $\mathbb{D}_2$. 
	
	Our key idea is to ``generate"
	a response $\tilde{Y}_{21}=V(X_{21},\eta)$. 
	Such a generated response $\tilde{Y}_{21}$ is generated through the generator function in (\ref{gen}) and evaluated at the covariate $X_{21}$
	in $\mathbb{D}_{21}$. 
	Then, the dataset consisting of the generated response and the corresponding covariates in $\mathbb{D}_{21}$ is denoted by $\widetilde{\mathbb{D}}_{21}=\{(\tilde{Y}_{21,i},X_{21,i})\}_{i=1}^{n_2/2}$. Note that
	\[
	\widetilde{\mathbb{D}}_{21}=\{(\tilde{Y}_{21,i},X_{21,i})\}_{i=1}^{n_2/2}\stackrel{\text { i.i.d }}{\sim} \mathbb{P}_{1,Y|X}\times\mathbb{P}_{2, X},
	\]
	and
	\[
	\mathbb{D}_{22}=\{(Y_{22,i},X_{22,i})\}_{i=1}^{n_2/2}\stackrel{\text { i.i.d }}{\sim}\mathbb{P}_{2,Y|X}\times\mathbb{P}_{2, X}.
	\]
	In such a way,  testing the equality of two conditional distributions $\mathbb{P}_{1,Y|X}=\mathbb{P}_{2,Y|X}$
	can be translated into the equality testing of their corresponding joint  distributions based on the two independent samples $\widetilde{\mathbb{D}}_{21}$ and $\mathbb{D}_{22}$.
	\begin{Remark}\label{full_alt}
		A natural question is: why not employ the generated dataset $\widetilde{\mathbb{D}}_2=\{(\tilde{Y}_{2,i},X_{2,i})\}_{i=1}^{n_2}$ together with the original dataset $\mathbb{D}_2=\{(Y_{2,i},X_{2,i})\}_{i=1}^{n_2}$ for two-sample testing, where $\tilde{Y}_{2,i}$ is generated via the generator function in (\ref{gen}) evaluated at the covariate $X_{2,i}$, thereby increasing the effective sample size? This approach is not considered in this work because $\widetilde{\mathbb{D}}_2$ and $\mathbb{D}_2$ exhibit substantial dependence owing to their shared data $\{X_{2,i}\}_{i=1}^{n_2}$, and the theoretical guarantees of many classical unconditional two-sample tests rely on sample independence. To avoid the technicality arising from highly correlated samples, we adopt the standard data splitting strategy, which has been considered in prior work \citep{shi2021double,zhang2025doubly}.
	\end{Remark}
	\begin{Remark}
		Another natural question is whether the conditional distribution equality in (\ref{cdt_problem}) could be assessed by conducting unconditional two-sample testing based  only on the response observations in $\widetilde{\mathbb{D}}_{21}$ and $\mathbb{D}_{22}$, namely $\{\tilde{Y}_{21,i}\}_{i=1}^{n_2/2}$ and $\{Y_{22,i}\}_{i=1}^{n_2/2}$. 
		We wish to note that this  is generally inapplicable for the following reason: even when $\tilde{Y}_{21}$ and $Y_{22}$ 
		have the same marginal distributions, and $X_{21}$ and $X_{22}$ are identically distributed,   the joint distributions of $(\tilde{Y}_{21},X_{21})$ and $(Y_{22},X_{22})$ may not necessarily be the same, a well-established result from optimal transport and copula theory \citep{villani2008optimal,jaworski2010copula}.
	\end{Remark}

	\subsection{The proposed procedure when the conditional generator is unknown}\label{unknown_cond_g} In practice,  complete information of $\mathbb{P}_{1,Y|X}$ is impossible. 
	Nonetheless,  in view of those state-of-the-art generative learning approaches such as conditional generative adversarial networks, MDNs etc,  a conditional generator for $\mathbb{P}_{1,Y|X}$ can be well estimated from the data $\mathbb{D}_1$, especially when the size of $\mathbb{D}_1$ is large. 
	
	In the following, we will introduce a general procedure for testing the equality of two conditional distributions. 
	Given $\mathbb{D}_1=\{(Y_{1,i},X_{1,i})\}_{i=1}^{n_1}$ and $\mathbb{D}_2=\{(Y_{2,i},X_{2,i})\}_{i=1}^{n_2}$, 
	let $A_\text{G}(\cdot)$ be 
	a conditional generative learning algorithm such as the MDNs or conditional GANs,  whose input is a dataset  and  output is  an estimated conditional generator for the corresponding conditional distribution. Meanwhile, let $A_{\text{TS}}(\cdot,\cdot,\cdot)$ be  a two-sample testing algorithm, 
	with triple inputs: the two datasets and a specified nominal size.  The output of $A_{\text{TS}}(\cdot,\cdot,\cdot)$ is a
	binary output $\{0,1\}$, where ``1'' signifies rejecting a null hypothesis. 
	
	Now, the rundown of our proposed  procedure for  testing the equality of two conditional distributions is as follows: 
	\begin{itemize}
		\item {\bf Step 1 (Conditional generative learning)}: Apply $A_\text{G}$ to the first data $\mathbb{D}_1$ and obtain the estimated conditional generative function $\widehat{V}$ 
		for the conditional distribution $\mathbb{P}_{1,Y|X}$. 
		\item {\bf Step 2 (Sample splitting and  synthetic response generation)}: Randomly divide the second data  $\mathbb{D}_2=\{(Y_{2,i},X_{2,i})\}_{i=1}^{n_2}$ into two equal-size sub-datasets
		\begin{equation}\label{data_split_D2}
			\mathbb{D}_{21}=\{(Y_{21,i},X_{21,i})\}_{i=1}^{n_2/2}\quad\text{and}\quad \mathbb{D}_{22}=\{(Y_{22,i},X_{22,i})\}_{i=1}^{n_2/2}.
		\end{equation}
		For $X_{21,i},i=1,2,\ldots,n_2/2$, generate/sample $\hat{Y}_{21,i}$ by $\widehat{V}$  and obtain a generated dataset
		\[
		\widehat{\mathbb{D}}_{21}=\{(\hat{Y}_{21,i},X_{21,i})\}_{i=1}^{n_2/2}.
		\]
		For example, let $\widehat{V}$ be an estimate of $V$  in (\ref{gen}).  One can generate the random noises $\eta_{1},\eta_{2},\ldots,\eta_{n_2/2}\stackrel{\text { i.i.d }}{\sim} \text{Unif}([0,1])$, then $\hat{Y}_{21,i}=\widehat{V}(X_{21,i},\eta_{i}), i=1,2,\ldots,n_2/2$.
		\item {\bf Step 3 (Two-sample unconditional distribution testing)}:  For the two datasets $\widehat{\mathbb{D}}_{21}$ and $\mathbb{D}_{22}$, apply a two-sample testing procedure $A_{\text{TS}}$ with a nominal size $\alpha\in (0,1)$ to conduct statistical testing for the hypothesis in (\ref{cdt_problem}).
	\end{itemize}
	
	In principle, one can apply our proposed framework  by incorporating any feasible conditional generative learning method and any existing two-sample testing approach. In this work,  we  consider conditional generative learning using mixture density networks and  propose the following generative classification-accuracy-based conditional distribution equality test (see Section \ref{CDET_CA}) under our new framework as the main example. 
	\section{One specific classification-based test with MDNs}\label{specific_test}
	In this section, to explicitly illustrate the test within our proposed framework, we propose a specific classification-based test using MDNs. In what follows, we first discuss about the conditional generative learning using MDNs and then present the specific proposed test.
	\subsection{Learning the unknown conditional generator using MDNs}\label{MDN_learner}
	
	Since a key component in our proposed testing  framework is to learn the conditional generator function,  
	in this subsection,  we will introduce a mixture density network model \citep[MDN,][]{bishop1994mixture} using deep neural networks to generate/sample the multivariate response.
	
	Aside from the notations in section \ref{CD-Testing},  more notations are needed. 
	Let $f_{k,Y|X}$ be the  conditional  p.d.f. of $\mathbb{P}_{k,Y|X}$, and $f_{k,X}$ be the p.d.f. of $\mathbb{P}_{k,X}$ for $k=1,2$.  Then, the joint p.d.f. $f_{k,Y,X}$ for the dataset $\mathbb{D}_k$ satisfies that $f_{k,Y,X}=f_{k,Y|X}f_{k,X}$, $k=1, 2$.  When there is no ambiguity, we also use  $f_k$ interchangeably to denote $f_{k,Y,X}$ for $k=1,2$.
	
	We first briefly review the feedforward neural networks (FNNs) that will be used.  A class of feedforward neural networks (FNNs) $\mathcal{F}$ consists of functions $F_{\boldsymbol{\phi}}:\mathbb{R}^{d_{\textup{in}}}\to\mathbb{R}^{d_{\textup{out}}}$ that is explicitly described by its input dimension $\textup{dim}_{\textup{in}}(\mathcal{F})=d_{\textup{in}}$, output dimension $\textup{dim}_{\textup{out}}(\mathcal{F})=d_{\textup{out}}$, weight and bias parameters $\boldsymbol{\phi}$, depth $\mathcal{D}$, width $\mathcal{W}$, size $\mathcal{S}$, number of neurons $\mathcal{U}.$  Specifically,
	\[
	F_{\boldsymbol{\phi}}(x)=
	\mathbb{A}_{\mathcal{D}}\circ\sigma_{a}\circ\mathbb{A}_{\mathcal{D}-1}\circ\sigma_{a}\circ\cdots \circ \sigma_{a} \circ \mathbb{A}_{1}\circ\sigma_{a}\circ\mathbb{A}_{0}(x),\ \ x\in\mathbb{R}^{d_{\textup{in}}},
	\]
	where $\mathbb{A}_{i}(z)=A_iz+b_i,z\in\mathbb{R}^{d_{i}}$ with weight matrix $A_i\in \mathbb{R}^{d_{i+1}\times d_i}$ and bias vector $b_i\in \mathbb{R}^{d_{i+1}},~i=0,1,\ldots,\mathcal{D}$, and $\sigma_{a}$ is the component-wise leaky rectified linear unit (Leaky-ReLU) activation function \citep{maas2013RectifierNI}, that is $\sigma_{a}(x)=x\mathbb{I}\{x>0\}+a x\mathbb{I}\{x\le0\}$ with $a\in[0,1)$ being a fixed parameter and $a=0$ corresponds to the widely-used ReLU activation function. Then, $d_0=d_{\textup{in}},d_{\mathcal{D}+1}=d_{\textup{out}}$ and $\boldsymbol{\phi}=(A_0,A_1,\ldots,A_{\mathcal{D}},b_0,b_1,\ldots,b_{\mathcal{D}})$.  And for this network, the width parameter
	$\mathcal{W}=\max\{d_i,i=1,\ldots,\mathcal{D}\}$ is the maximum width of hidden layers; the number of neurons $\mathcal{U}$ is defined as the number of neurons in $F_{\boldsymbol{\phi}}$, i.e., $\mathcal{U}=\sum_{i=1}^{\mathcal{H}}d_i$; the size $\mathcal{S}$ is the total number of parameters in the network. 
	
	We propose to estimate the conditional density $f_{1,Y|X}$ using  the following multivariate conditional mixture density network model: 
	\[
	f_{G}(y,x|\btheta)=\sum_{g=1}^G \frac{\alpha_g(x;\btheta)}{(2 \pi)^{\frac{p}{2}}\sigma_g^{p}(x;\btheta)}  \exp\left\{-\frac{\|y-\mu_g(x;\btheta)\|_2^2}{2\sigma_g^{2}(x;\btheta)}\right\},
	\]
	where $(\alpha_g,\mu_g,\sigma_g,g=1,2,\ldots,G)$ satisfies that $\sum_{g=1}^{G}\alpha_g=1,\alpha_g\ge0,\sigma_g>0$ for $g=1,2,\ldots,G$ and is expressed by a multi-output FNN parametrized by $\btheta$. That is, there exists a FNN function $F_{\btheta}$, such that $F_{\btheta}(x)=(\alpha_g(x;\btheta),\mu_g(x;\btheta),\sigma_g(x;\btheta),g=1,2,\ldots,G)$ for any $x$.  
	\begin{Remark}
		The MDNs we adopt  is a slightly modified version of the classical MDNs in the following sense: (1) we use the 
		ReLU activation function in MDNs to mitigate the gradient vanishing problem,  rather than the sigmoidal activation function in the classical MDNs; 
		(2) to handle the multivariate response,  \cite{zhou2023testing} applied the chaining rule to the probability density function  and turned the multivariate generative learning problem into a sequence of univariate generative learning tasks.  Instead, 
		we employ MDNs for direct multivariate generative learning, which may circumvent the potential  deteriorating data generation problem in solving a series of deep generative learning problems sequentially.
	\end{Remark}
	
	The empirical objective function for the mixture density
	network model is the empirical log-likelihood given by
	\[
	\widehat{\mathbb{L}}_{n_1}(\mathbb{D}_1;\btheta)=\frac{1}{n_1}\sum_{i=1}^{n_1}\log f_{G}(Y_{1,i},X_{1,i}|\btheta). 
	\]
	Define 
	\[
	\hat{\btheta}_{n_1} \in \mathop{\arg}\mathop{\max}\limits_{\btheta\in\Theta_{\text{mix}}}\widehat{\mathbb{L}}_{n_1}(\mathbb{D}_1;\btheta),
	\]
	where $\Theta_{\text{mix}}$ is the network parameter space.
	An alternative way to write the neural network function class is to define 
	\begin{equation}\label{F_mix}
		\mathcal{F}_{\text{mix}}=\{F_{\btheta}:\btheta\in\Theta_{\text{mix}}\}. 
	\end{equation}
	That is
	$\mathcal{F}_{\text{mix}}$  is a class of FNNs with parameter $\btheta$, $\textup{dim}_{\textup{in}}(\mathcal{F}_{\text{mix}})
	=d,\textup{dim}_{\textup{out}}(\mathcal{F}_{\text{mix}})=G(p+2)$, depth
	$\mathcal{D}_{\mathcal{F}_{\text{mix}}}$ width $\mathcal{W}_{\mathcal{F}_{\text{mix}}}$, and  size $\mathcal{S}_{\mathcal{F}_{\text{mix}}}$.  Note that the parameters of $\mathcal{F}_{\text{mix}}$ 
	might depend on the sample size $n_1$, but we suppress this dependence for notational simplicity. 
	Then,  the resulting estimator for $f_{1,Y|X}$ is  defined as 
	\begin{equation}\label{mdn_est_cdensity}
		\hat{f}_{1,Y|X}(y,x)=f_{G}(y,x|\hat{\btheta}_{n_1}).
	\end{equation}
	We summarize the conditional density estimation  along with the associated  generating/sampling procedure in Algorithm \ref{Training_ConGL}.
	
	Note that step 5 -- step 7 in  Algorithm \ref{Training_ConGL} are the conditional generating/sampling procedure based on the estimated conditional density  $\hat{f}_{1,Y|X}$, attributed to the Gaussian mixture nature. 
	We denote the distribution of the generated  $\hat{Y}$ given $X=x$  in Algorithm \ref{Training_ConGL} by $\hat{\mathbb{P}}_{1,Y|X=x}$, 
	whose density function is $\hat{f}_{1,Y|X}$. Thus,  when there is no confusion, 
	we  sometimes  refer the conditional density estimation   as the  conditional generator learning.   
	
	Given the dataset $\mathbb{D}_{21}=\{(Y_{21,i},X_{21,i})\}_{i=1}^{n_2/2}$,  we can obtain the generated dataset $\widehat{\mathbb{D}}_{21}=\{(\hat{Y}_{21,i},X_{21,i})\}_{i=1}^{n_2/2}$ according to the sampling procedure in Algorithm \ref{Training_ConGL}.
	
	\begin{algorithm}[h]
		\caption{Conditional generative learning using MDNs}\label{Training_ConGL}
		\hspace*{0.02in} {\bf Require:}
		Data $\{(X_{1,i},Y_{1,i})\}_{i=1}^{n_1}$, number of mixture Gaussian distributions: $G$, batch size $m$, and a unlabeled predictor data point $x$.
		\begin{algorithmic}[1]
			\While{not converged} 
			\State Draw $m$ minibatch samples $\{(X_{1,bi},Y_{1,bi})\}_{i=1}^{m}$ from $\{(X_{1,i},Y_{1,i})\}_{i=1}^{n_1}$.
			\State Update the Mixture Density Network $f_{\btheta}$ by descending its stochastic gradient:
			\[
			\nabla_{\btheta}\left[\frac{1}{m}\sum_{i=1}^{m}\sum_{g=1}^G\frac{\alpha_g(X_{1,bi};\btheta)}{(2 \pi)^{\frac{p}{2}}\sigma_g^{p}(X_{1,bi};\btheta)}  \exp\left\{-\frac{\|Y_{1,bi}-\mu_g(X_{1,bi};\btheta)\|_2^2}{2\sigma_g^{2}(X_{1,bi};\btheta)}\right\}\right].
			\]
			\EndWhile
			\State  Denote $\hat{\balpha}_{n_{1}}=(\alpha_{1}(\cdot;\hat{\btheta}_{n_1}),\ldots, \alpha_{G}(\cdot;\hat{\btheta}_{n_1}))$, $\hat{\bmu}_{n_{1}}=(\mu_{1}(\cdot;\hat{\btheta}_{n_1}),\ldots, \mu_{G}(\cdot;\hat{\btheta}_{n_1}))$, $\hat{\bsigma}_{n_{1}}=(\sigma_{1}(\cdot;\hat{\btheta}_{n_1}),\ldots, \sigma_{G}(\cdot;\hat{\btheta}_{n_1}))$, where $\hat{\btheta}_{n_1}$ is the ultimate estimate of the network weight and bias parameters.
			\State Let $g_v\in\{1,2\ldots,G\}$ be an integer sampled from a discrete distribution satisfying $\mathbb{P}(I_v=g)=\alpha_{g}(x;\hat{\btheta}_{n_1})$ for $g=1,\ldots, G$, where $I_v$ is a random variable.
			\State Randomly generate a $p$-dimensional vector $W$ from the $p$-dimensional standard normal distribution;
			\State \Return $\hat{Y}$, where $\hat{Y}=\mu_{g_v}(x;\hat{\btheta}_{n_1})+\sigma_{g_v}(x;\hat{\btheta}_{n_1})W$.
		\end{algorithmic}
	\end{algorithm}
	
	\subsection{The generative classification-accuracy-based conditional distribution equality test}\label{CDET_CA}
	
	In this subsection, we propose the generative classification accuracy-based conditional distribution equality test (GCA-CDET).  
	Given  two independent random samples, 
	the main idea of the  classification accuracy-based unconditional two-sample test is to treat the two-sample testing problem  as a binary classification problem. 
	With this view,  we can introduce the proposed GCA-CDET below.

	First, we apply steps 1-4 in Algorithm \ref{Training_ConGL} to  $\mathbb{D}_1$ and obtain the estimated conditional density $\hat{f}_{1,Y|X}$. Applying data splitting to $\mathbb{D}_2$ as  in (\ref{data_split_D2}), we get 
	\[
	\mathbb{D}_{21}=\{(Y_{21,i},X_{21,i})\}_{i=1}^{n_2/2}\quad\text{and}\quad \mathbb{D}_{22}=\{(Y_{22,i},X_{22,i})\}_{i=1}^{n_2/2}.
	\]
	Then, according to steps 5-7 in Algorithm \ref{Training_ConGL} to $\mathbb{D}_{21}$,  we can obtain the generated dataset $\widehat{\mathbb{D}}_{21}=\{(\hat{Y}_{21,i},X_{21,i})\}_{i=1}^{n_2/2}$.
	
	Once the datasets $\widehat{\mathbb{D}}_{21}$ and $\mathbb{D}_{22}$ are ready, 
	we treat $\widehat{\mathbb{D}}_{21}$ as data from class ``1'' and randomly split $\widehat{\mathbb{D}}_{21}$ into two equal-size subsets
	\[
	\widehat{\mathbb{D}}_{211}=\{(\hat{Y}_{211,i},X_{211,i})\}_{i=1}^{n_2/4}~\text{and}~\widehat{\mathbb{D}}_{212}=\{(\hat{Y}_{212,i},X_{212,i})\}_{i=1}^{n_2/4};
	\]
	likewise, we treat $\mathbb{D}_{22}=\{(Y_{22,i},X_{22,i})\}_{i=1}^{n_2/2}$ as data from class ``0'' and randomly partition it into two equal-size subsets
	\[
	\mathbb{D}_{221}=\{(Y_{221,i},X_{221,i})\}_{i=1}^{n_2/4}~\text{and}~\mathbb{D}_{222}=\{(Y_{222,i},X_{222,i})\}_{i=1}^{n_2/4}.
	\]
	We use $\widehat{\mathbb{D}}_{211}$ and $\mathbb{D}_{221}$ to train a classifier based on nonparametric logistic regression using neural networks. Define the pooled data 
	$\{({Y}_{\text{po},i}, {X}_{\text{po},i}, {S}_{\text{po},i})\}_{i=1}^{n_2/2}=\{(\hat{Y}_{211,i},X_{211,i},1)\}_{i=1}^{n_2/4}\cup \{(Y_{221,i},X_{221,i},0)\}_{i=1}^{n_2/4}$. 
	The logistic classification loss function is 
	\[
	\widehat{\mathbb{L}}_{acc}(\widehat{\mathbb{D}}_{211},\mathbb{D}_{221};R)=\frac{2}{n_2}\sum_{i=1}^{n_2/2}\ell(R, {Y}_{\text{po},i}, {X}_{\text{po},i}, {S}_{\text{po},i}),
	\]
	where  $\ell(R, y, x, s)=-s R(y,x)+\log \left(1+e^{R(y,x)}\right)$. 
	Define 
	\begin{equation}
		\label{classifier_def}
		\hat{R}_{n_2}\in \argmin_{R\in\mathcal{R}}\widehat{\mathbb{L}}_{acc}(\widehat{\mathbb{D}}_{211},\mathbb{D}_{221};R), 
	\end{equation}
	where  $\mathcal{R}$ is a FNN. 
	The resulting  classifier is given by
	$\hat{C}_{n_2}(y,x)=\mathbb{I}(\hat{R}_{n_2}(y,x)\ge1).$
	Based on $\hat{C}_{n_2}(y,x)$, we calculate the classification errors on $\widehat{\mathbb{D}}_{212}$ and $\mathbb{D}_{222}$:
	\[
	\hat{e}_1=\frac{4}{n_2}\sum_{(y,x)\in\widehat{\mathbb{D}}_{212}}\mathbb{I}\{\hat{C}_{n_2}(y,x)\neq 1\}\ \ \text{ and }\ \ \hat{e}_0=\frac{4}{n_2}\sum_{(y,x)\in\mathbb{D}_{222}}\mathbb{I}\{\hat{C}_{n_2}(y,x)\neq 0\},
	\]
	respectively. Intuitively, under $H_0: \mathbb{P}_{1,Y|X}=\mathbb{P}_{2,Y|X}$,  if the generator is well learned, the well-trained classifier would be very closed to random guess and hence $\hat{e}_1\approx\hat{e}_0\approx1/2$. Hence, based on the central limit theorem and given a significance level $\alpha\in(0,1)$, we reject the null hypothesis when $\phi_{\text{acc},\alpha}(\mathbb{D}_1,\mathbb{D}_2)=1$, where
	\[
	\phi_{\text{acc},\alpha}(\mathbb{D}_1,\mathbb{D}_2)=\mathbb{I}\left\{\frac{\hat{e}_1+\hat{e}_0-1}{\sqrt{\hat{e}_1(1-\hat{e}_1)/(n_2/4)+\hat{e}_0(1-\hat{e}_0)/(n_2/4)}}<-z_\alpha\right\},
	\]
	and $z_\alpha$ is the upper $(1-\alpha)$ quantile of the standard Gaussian distribution.
	\section{Theoretical properties}\label{theory}
	In this section, we  investigate the properties of the learned MDN conditional generator and the proposed specific test GCA-CDET. Additionally, we  provide the detailed analysis of the minimax optimality of the proposed framework in Section S.4 of the online supplementary material.
	
	\subsection{Convergence rates of the MDN conditional generator}
	We first provide theoretical analysis of the MDNs-based conditional generative learning.  We 
	will study  the convergence properties of the estimated $\hat{f}_{1,Y|X}$ defined in (\ref{mdn_est_cdensity}), which can imply 
	the convergence properties of $\hat{\mathbb{P}}_{1,Y|X}$ under certain distribution distance. 
	
	Some regularity conditions of the target distribution are needed.  We adopt the H\"older density functions in this work.
	Without loss of generality$^*\footnote{
		Although we assume bounded supports for all related distributions in the current work, analogous results can be derived by positing specific tail decay rates for distributions on some unbounded supports, with exceptions on those minimax optimality results. However this is not essential since such a relaxation is at the price of additional logarithmic terms and will make the result unnecessarily complicated, and hence we omit it for clarity.}$, we assume that $\mathcal{Y}=[0,1]^p$ and $\mathcal{X}=[0,1]^d$. 
	We now give a definition of  H\"older functions. 
	\begin{definition}[H\"older  class]\label{HolderClass}
		A H\"older class $\mathcal{H}^\beta([0,1]^d,M)$ with $\beta=k+a$ where $k\in \mathbb{N}^+$ and $a\in(0,1]$
		consists of function $f:[0,1]^d\to\mathbb{R}$ satisfying
		\[
		\max\limits_{\|\boldsymbol{\alpha}\|_1\le k}\|\partial^{\boldsymbol{\alpha}}f\|_{\infty}\le M,\max\limits_{\|\boldsymbol{\alpha}\|_1= k}\max\limits_{x\neq y}\frac{|\partial^{\boldsymbol{\alpha}}f(x)-\partial^{\boldsymbol{\alpha}}f(y)|}{\|x-y\|_2^a}\le M,
		\]
		where $\|\boldsymbol{\alpha}\|_1=\sum_{i=1}^{d}\alpha_i$ and  $\partial^{\boldsymbol{\alpha}}=\partial^{\alpha_1}\partial^{\alpha_2}\cdots\partial^{\alpha_d}$ for $\boldsymbol{\alpha}=(\alpha_1,\alpha_2,\ldots,\alpha_d)\in \mathbb{N}^{+d}$.
	\end{definition}
	We next define a distribution density class
	\begin{equation}\label{density_class_def_holder}
		\begin{split}
			\mathcal{H}_{M,\beta,c_1,c_2}&=\left\{p_{Y, X}(y, x):p_{Y|X}(y|x)\in\mathcal{H}^\beta([0,1]^{p+d},M),p_{X}(x)\in\mathcal{H}^\beta([0,1]^{d},M), \right.\\
			&\hspace{0.8cm} \left. c_1\le \inf_{y,x}p_{Y|X}(y, x)\wedge \inf_{x}p_{X}(x)\le \sup_{y,x}p_{Y|X}(y, x)\vee \sup_{x}p_{X}(x)\le c_2\right\},
		\end{split}
	\end{equation}
	where $c_1,c_2$ are two positive constants satisfying $c_1<1<c_2$, $\beta\ge1$ and $M>0$. In (\ref{density_class_def_holder}), $p_{Y, X}(y, x)=p_{Y|X}(y|x)p_{X}(x)$ is a joint p.d.f. of some random variable pair $(Y,X)$ on $[0,1]^p\times[0,1]^d$, $p_{Y|X}(y|x)$ is the corresponding conditional p.d.f. of $Y$ given $X$ and $p_{X}(x)$ is the marginal p.d.f. of $X$. To lighten the notation, we may  use $\mathcal{H}$ to denote $\mathcal{H}_{M,\beta,c_1,c_2}$ in the subsequent discussions, unless there is any potential ambiguity.
	
	The following two assumptions are imposed.
	
	\begin{assumption}\label{f1_joint_holder_cond_general}
		The joint p.d.f. $f_{1,Y,X}\in \mathcal{H}_{M,\beta,c_1,c_2}$.
	\end{assumption}
	\begin{assumption}\label{F_net_cond_general}
		The function class $\mathcal{F}_{\text{mix}}$ in (\ref{F_mix}) is a ReLU-activated FNN and has depth $\mathcal{D}_{\mathcal{F}_{\text{mix}}}=21L\lceil\log_2(8L)\rceil(\lfloor\beta \rfloor+1)^2+2(p+d)$ and width $\mathcal{W}_{\mathcal{F}_{\text{mix}}}=38(\lfloor\beta \rfloor+1)^2 (p+d)^{\lfloor\beta \rfloor+1} 3^{p+d}(p+2)GN\lceil\log_2(8N) \rceil$ with $NL\asymp (n_1G)^{\frac{d}{2(2\beta+d)}},\quad G^{2+\frac{2}{p(p+2)}}\asymp (NL)^{\frac{4\beta}{d}}$.
		In addition, for any $\btheta\in\Theta_{\text{mix}}$, which induces $\mathcal{F}_{\text{mix}}$ as in (\ref{F_mix}), it holds that $c_1\le\inf_{y,x}f_{G}(y,x|\btheta)\le \sup_{y,x}f_{G}(y,x|\btheta)\le c_2+C_2$ and $\inf_{x}\sigma_g(x;\btheta)\ge C_1G^{-1/\{p(p+2)\}},g\in[G]$, where $C_1,C_2$ are two constants defined in Lemma B.4.
	\end{assumption}
	
	The next  theorem establishes a $L_1$-bound for  $\hat{f}_{1,Y|X}$ defined in (\ref{mdn_est_cdensity}).
	\begin{theorem}[Nonasymptotic upper bound of the MDNs-based conditional density estimator]\label{mdn_generator_rates}
		Under Assumptions \ref{f1_joint_holder_cond_general} \& \ref{F_net_cond_general}, it holds that
		\begin{equation}\label{mdn_l1_bound}
			\mathbb{E}_{\mathbb{D}_1}\|f_{1,Y|X}-\hat{f}_{1,Y|X}\|_{1}\le Cn_1^{-\frac{2\beta}{c_{p}(\beta+d)}}\log^\frac{7}{2} n_1,
		\end{equation}
		where $c_{p}=2p^2+4p+4$ and $C$ is an absolute constant depending on $\beta,c_1,c_2,M,p,d$.
	\end{theorem}
	The $L_1$-bound of density functions is closely related to the total variation distance.  Our proof of Theorem \ref{mdn_generator_rates} rests on the recently-developed offset Rademacher complexity \citep{liang2015learning}. This is different from the proof of MDN in \cite{zhou2023testing}, where they employed the classic localization technique  \citep{farrell2021deep}.  
	One of our technical contributions in this paper is to 
	derive a new empirical process bound incorporating the offset Rademacher complexity, which 
	can significantly simplify the proof related to neural networks and could be of independent interest.  For more details, we refer the readers to Lemma S.7 in Section S.5 of the supplementary material, where we provide the detailed offset Rademacher complexity inequality and its related discussions.  Further discussions on the rate difference between the convergence rate in (\ref{mdn_l1_bound}) and those by \cite{zhou2023testing} are provided in Section S.2 of the online supplementary material.

	\begin{Remark}
		The convergence rate in Theorem \ref{mdn_generator_rates} may be extremely slow when $d$ is large and hence demonstrates the curse of dimensionality. In Section S.3 of the online supplementary material, to mitigate the curse of dimensionality in learning conditional generator, we derive an approximation result using neural networks (Lemma S.1) and an improved convergence rate for the learned conditional generator under a low-dimensional sufficient representation assumption (Theorem S.2).
	\end{Remark}

	For two distributions $\mathbb{P}$ and $\mathbb{Q}$ with densities $p(\cdot)$ and $q(\cdot)$ on a measurable space $\mathcal{Z}$, the total variation distance between $\mathbb{P}$ and $\mathbb{Q}$ is defined as $\text{TV}(\mathbb{P},\mathbb{Q}):=\sup_{B\subset \mathcal{Z},B\text{ measurable}}|\mathbb{P}(B)-\mathbb{Q}(B)|.$
	It is well known that 
	$\text{TV}(\mathbb{P},\mathbb{Q})=({1}/{2})\int_{\mathcal{Z}}|p(z)-q(z)|d\mu(z)=({1}/{2})\|p-q\|_1;$
	see (15.6) of \cite{wainwright2019high} for an example. In view of the relationship between the total variation distance and $L_1$ norm, 
	the next corollary is a direct consequence of  Theorem \ref{mdn_generator_rates}.
	\begin{corollary}[Nonasymptotic upper bound of the conditional generator in total variation distance]\label{mdn_generator_rates_tv}
		Under Assumptions \ref{f1_joint_holder_cond_general} \& \ref{F_net_cond_general},  we have
		\[
		\mathbb{E}_{\mathbb{D}_1}\mathbb{E}_{X'\sim\mathbb{P}_{2,X}}\textnormal{TV}(\hat{\mathbb{P}}_{1,Y|X=X'},\mathbb{P}_{1,Y|X=X'})\le C_1n_1^{-\frac{2\beta}{c_{p}(\beta+d)}}\log^\frac{7}{2} n_1
		\]
		where $C_1=c_2C/2$,  $c_{p}$ and $C$ are the absolute constants defined in Theorem \ref{mdn_generator_rates}.
	\end{corollary}
	
	\subsection{Consistency of GCA-CDET}
	In this subsection, we will prove the consistency of the proposed GCA-CDET.
	To this end, additional conditions are needed. 
	\begin{assumption}\label{conds_on_sample_sizes}
		Assume that $n_2n_1^{-\omega_0}\to0$ as $n_2\to\infty$, where $\omega_0=(2\beta-\delta)/(c_{p}(\beta+d))$ for some $\delta\in(0,2\beta)$.
	\end{assumption}
	\begin{assumption}\label{classification_nets}
		The function class $\mathcal{R}$ in (\ref{classifier_def}) is a ReLU neural network and has depth $\mathcal{D}_{\mathcal{R}}=21L\lceil\log_2(8L)\rceil(\lfloor\beta \rfloor+1)^2+2(p+d)$ and width $\mathcal{W}_{\mathcal{R}}=38(\lfloor\beta \rfloor+1)^2 (p+d)^{\lfloor\beta \rfloor+1} 3^{p+d}N\lceil\log_2(8N) \rceil$ with $NL\asymp (n_2)^{\frac{d}{2(2\beta+d)}}$.
	\end{assumption}
	\begin{assumption}\label{hypothesis_assumps}
		There exists a constant $\varepsilon>0$ such that $(f_{1,Y,X},f_{2,Y,X}) \in \mathcal{P}_0$ under the null hypothesis and $(f_{1,Y,X},f_{2,Y,X}) \in \mathcal{P}_1(\varepsilon)$ under the alternative hypothesis, where $\mathcal{P}_0$ and $\mathcal{P}_1(\varepsilon)$ are defined as below:
		\begin{equation*}
			\mathcal{P}_0=\{(f_{1,Y,X},f_{2,Y,X}):f_{i,Y,X}\in\mathcal{H},i=1,2,~f_{1,Y|X}=f_{2,Y|X}\},
		\end{equation*}
		\begin{equation*}
			\mathcal{P}_1(\varepsilon)=\{(f_{1,Y,X},f_{2,Y,X}):f_{i,Y,X}\in\mathcal{H},i=1,2,~\|f_{1,Y|X}-f_{2,Y|X}\|_2\ge\varepsilon \}.
		\end{equation*}
	\end{assumption}
	
	Assumption \ref{conds_on_sample_sizes} ensures that there are sufficient samples for the MDN to obtain a satisfactory conditional density estimator.  Assumption \ref{classification_nets} is needed to ensure the classifier can be well trained.  Under Assumption \ref{classification_nets}, the estimated $\hat{R}_{n_2}$ defined in (\ref{classifier_def}), which is instrumental in obtaining the learned classifier, 
	achieves the minimax rate in classical nonparametric regression  \citep{stone1982optimal}; see (88) in the proof of Theorem \ref{ca_test_consistency} for more details. Similar assumptions can be found in \cite{jiao2023deep} for  nonparametric regression problem.
	Assumption \ref{hypothesis_assumps} signifies the distinct separation of the conditional distributions of the two datasets under the alternative hypothesis.
	
	\begin{theorem}[Testing consistency of GCA-CDET]
		\label{ca_test_consistency}
		Suppose that Assumptions \ref{F_net_cond_general}, \ref{conds_on_sample_sizes}, \ref{classification_nets} \& \ref{hypothesis_assumps} hold. 
		Then, 
		\begin{itemize}
			\item[(1)] Under the null hypothesis $H_0: \mathbb{P}_{1,Y|X}=\mathbb{P}_{2,Y|X}$, $\lim_{n_2\to\infty}\mathbb{E}_{H_0}\phi_{\text{acc},\alpha}(\mathbb{D}_1,\mathbb{D}_2) \leq \alpha$.
			\item[(2)] Under the alternative hypothesis $H_1: \mathbb{P}_{1,Y|X}\neq\mathbb{P}_{2,Y|X}$, the asymptotic test is consistent as $\lim_{n_2\to\infty}\mathbb{E}_{H_1}\phi_{\text{acc},\alpha}(\mathbb{D}_1,\mathbb{D}_2)=1$.
		\end{itemize}
	\end{theorem}
	Theorem \ref{ca_test_consistency} tells that under some mild conditions, asymptotically, under the null hypothesis, the proposed  generative classification accuracy-based  testing approach can well control the type-I error; and under the alternative hypothesis, the power  tends to 1 as the sample size increases.

	\section{Simulation studies}\label{sec:simu}
	\subsection{Simulation}
	In this part, we conduct simulation studies to illustrate the performance of our proposed testing methodology. 
	We implement the proposed GCA-CDET and its oracle version  with $\mathbb{P}_{1,Y|X}$ known.  For comparison, we consider the Generative Conditional Independence Test \citep[GCIT,][]{bellot2019conditional}, the Doubly Robust Generative Conditional Independence
	Test \citep[DRGCIT,][]{zhang2025doubly}, and the Weighted Conformal Prediction based Test  \citep[WCPT,][]{hu2023two}.
	To evaluate the robustness of the proposed GCA-CDET, two approaches are utilized for training the classifier under both oracle and non-oracle settings: a nonparametric logistic regression implemented via neural networks (denoted as NN) and a linear logistic regression (denoted as LLR).
	For a fair comparison, all methods employ two-layer fully connected neural networks with the same width. Architectural details are provided in Section S.9 of the online supplementary materials.  It is worthy to note that although GCIT and DRGCIT may demonstrate inferior performance relative to the proposed GCA-CDET in certain simulation scenarios and real data analyses presented subsequently, this does not impact the conclusions regarding their strong performance in conditional independence testing. This is due to the fact that these methods were not specifically designed for the conditional distribution equality testing problem, and it is well-established that no statistical method achieves universal superiority across all scenarios.

	For the simulation study, six models are considered, each of which generates two datasets,  denoted by $\mathbb{D}_1$ and $\mathbb{D}_2$. 
	The covariates and response in each dataset $\mathbb{D}_k$ ($k=1,2$) are denoted by $X_{k}=(X_{k,1},\ldots,X_{k,d})\in\mathbb{R}^{d}$ and $Y_{k}=(Y_{k,1},\ldots,Y_{k,p})\in\mathbb{R}^{p}$, respectively.
	In Models M1-M6, the noise terms $\epsilon_k$'s follow the standard normal distribution and are independent of $X_{k}$ for $k=1,2$.

	\noindent {\bf M1. } (Uniform mixture, linear, homogeneous) {\it Let $Y_{k}=\alpha_{k}+\beta^{\top}X_{k}+\epsilon_{k},$ 
		where $X_{1}\sim $ $N\left(0, I_5\right)$, $X_{2}\sim \rm{Unif}\left(([-1.0,-0.5]\cup[0.5,1.0])^{5}\right)$, and $\beta=(1,-1,1,-1,1)^{\top}$.}
	
	\noindent {\bf M2. }  (Gaussian, linear, homogeneous) {\it Let $Y_{k}=\alpha_{k}+\beta^{\top}X_{k}+\epsilon_{k},$ 
		where $X_{1}\sim N\left(0, I_5\right)$ and $X_{2}\sim N\left(\mu, I_5\right)$ with $\mu=(1,1,-1,-1,0)^{\top}$, and $\beta=(1,-1,1,-1,1)^{\top}$.}
	
	\noindent {\bf M3.} \nolinebreak\hspace*{0.08cm}(Uniform mixture, nonlinear, homogeneous)
	{\it Let $Y_{k}=\alpha_{k}+\exp(X_{k,1}/2+X_{k,2}/2)-X_{k,3}\sin(X_{k,4}+X_{k,5})+\epsilon_{k},$ 
		where $X_{1}\sim N\left(0, I_5\right)$, $X_{2}\sim \rm{Unif}\left(([-1.0,-0.5]\cup[0.5,1.0])^{5}\right)$. }
	\noindent {\bf M4. } (Uniform mixture, nonlinear, heteroskedastic)
	{\it Let $Y_{k}=\alpha_{k} + X^2_{k,1} + \exp(X_{k,2}+X_{k,3}/3) + X_{k,4} - X_{k,5} + (0.5+X^2_{k,6}/2+X^2_{k,7}/2)\epsilon_{k}, $
		where $X_{1}\sim N\left(0, I_{10}\right)$, and $X_{2}\sim \rm{Unif}\left(([-1.0,-0.5]\cup[0.5,1.0])^{10}\right)$.}

	\noindent {\bf M5. } (Uniform mixture, nonlinear, homogeneous, high-dimensional)
	{\it Let $Y_{k}=\alpha_{k}+\exp(X_{k,1}/2+X_{k,2}/2)-X_{k,3}\sin(X_{k,4}+X_{k,5})+\epsilon_{k},$ 
		where $X_{1}\sim N\left(0, I_{100}\right)$ and $X_{2}\sim \rm{Unif}\left(([-1.0,-0.5]\cup[0.5,1.0])^{100}\right)$.  }

	\noindent {\bf M6. } (Bivariate response)
	{\it Let }$Y_{k,1}\! =\! \alpha_k \!+\! \beta^{\top}X_{k} \!+\!({u_{k}}/{2\pi})\sin(2u_k) \!+ \!\epsilon_{k,1}, $ $Y_{k,2} \!=\! \alpha_k \!+\! \beta^{\top}X_{k}\! +$\\$ ({u_{k}}/{2\pi})\cos(2u_k) + \epsilon_{k,2},$  
	{\it where} $X_{1}\sim N\left(0, I_5\right)$,  
	$X_{2}\sim \rm{Unif}\left(([-1.0,-0.5]\cup[0.5,1.0])^{5}\right)$, $u_k\sim \rm{Unif}[0,2\pi]$, $\epsilon_{k,l}\sim N(0,0.1^2),$ {\it and} $\epsilon_{k,l}\perp \!\!\! \perp X_k$ {\it for} $k,l=1,2$.

	To set the null  and alternative hypotheses, $\alpha_1$ and $\alpha_2$ are specified as follows: under the null,  $\alpha_1=\alpha_2=0$;  under the alternative, $\alpha_1=0$ and $\alpha_2=0.5$. 
	We set $n_1=n_2\in\{1000,2000\}$. 
	The nominal significance level is $\alpha=0.05$. 
	Each setup is repeated 500 times, with the results summarized in Table \ref{tab:simulation result} and the average running times reported in Table \ref{tab:time}.

	Table \ref{tab:simulation result} shows that both the type I error and statistical power of the Oracle method (with known $P_{1,Y|X}$) are close to the nominal level across all settings, thereby validating the motivation of the proposed framework.
	Across all settings, the proposed GCA-CDET method achieves empirical type I errors and powers that are comparable to, or closer to, those by the Oracle method, highlighting its superior performance.
	In nearly all cases, the proposed method also outperforms WCPT in controlling type I error.
	Although GCIT performs comparably to GCA-CDET in terms of the type I error control, it exhibits low power against the alternative.
	The proposed GCA-CDET performs comparable to DRGCIT in M1, M3 and M4, and slightly better in M5 and M6. But the proposed method takes less computation time compared to DRGCIT, as shown in Table \ref{tab:time}.
	Overall,  our proposed GCA-CDET is  a competitive method  for  two-sample conditional distribution equality testing.

	\begin{table}[h]{\color{black}
			\centering
			\caption{Percentage of rejections over 500 repetitions in the simulation studies.}\label{tab:simulation result}
			\begin{threeparttable}
				\resizebox{\textwidth}{!}{\begin{minipage}{\textwidth}
						\begin{tabular}{ccc c cc c cc cc c cc c  cc c cc c cc c}
							\hline
							& & &\multicolumn{8}{c}{Null} & \multicolumn{8}{c}{Alternative} \\
							\cline{4-11} \cline{13-20}
							& &  &\multicolumn{2}{c}{Oracle} &&   \multicolumn{2}{c}{GCA-CDET} & \multirow{2}{*}{GCIT}&  \multirow{2}{*}{DRGCIT}& \multirow{2}{*}{WCPT} && \multicolumn{2}{c}{Oracle} &&   \multicolumn{2}{c}{GCA-CDET} &\multirow{2}{*}{GCIT}&  \multirow{2}{*}{DRGCIT} &  \multirow{2}{*}{WCPT}  \\ 
							\cline{4-5} \cline{7-8} \cline{13-14} \cline{16-17}
							& $n_1$ & $n_2$ & NN & LLR && NN & LLR &  &  &  &  &  NN & LLR && NN & LLR && &   \\
							\hline
							\multirow{2}{*}{M1} & 1000 & 1000 & 0.050 & 0.048 && 0.050 & 0.048 & 0.064 &  0.078 & 0.518 &&   0.966 & 0.994 && 0.960 & 0.994 & 0.532 & 0.986 & 1.000  \\
							& 2000 & 2000  & 0.052 & 0.056 && 0.056 & 0.050 & 0.058 & 0.064 &  0.470 & & 1.000 & 1.000 && 1.000 & 1.000 & 0.536 &  1.000 & 0.982  \\
							\\
							\multirow{2}{*}{M2} & 1000 & 1000 & 0.050 & 0.046 && 0.046 & 0.064 & 0.074 & 0.566 & 0.060 && 0.956 & 0.992 && 0.826& 0.918 & 0.594& 1.000 & 0.768  \\
							& 2000 & 2000 & 0.050 & 0.050 & & 0.068 & 0.062 & 0.064 & 0.570 & 0.054    && 1.000 & 1.000 && 0.986 & 1.000 &0.612 & 1.000 &0.956  \\
							\\  \multirow{2}{*}{M3} & 1000 & 1000  & 0.046 & 0.046 && 0.060 & 0.042 & 0.070 & 0.086  & 0.212 &  &0.958 & 0.956 && 0.958 & 0.962 & 0.360&  1.000& 1.000 \\
							& 2000 & 2000  & 0.052 & 0.064 && 0.066 & 0.056 & 0.068 & 0.080 & 0.162   && 1.000 & 1.000 &&  1.000 & 1.000 & 0.348 & 1.000& 0.984   \\
							\\  \multirow{2}{*}{M4} & 1000 & 1000  & 0.044 & 0.056 && 0.076 & 0.074 & 0.072 &  0.050 & 0.994 &&  0.934 & 0.938 && 0.932 & 0.946 & 0.352 & 0.982 &0.988  \\
							& 2000 & 2000  & 0.042 & 0.056 && 0.064 & 0.072 & 0.066 & 0.048  & 1.000  && 0.996 & 1.000 &&  0.998 & 0.998 &0.356 & 1.000 &0.998    \\
							\\  \multirow{2}{*}{M5} & 1000 & 1000  & 0.052 & 0.058 && 0.064 & 0.068 & 0.092 & 0.108 & 0.548& &0.988 & 1.000 && 1.000 & 1.000 & 0.542 & 1.000 &1.000 \\
							& 2000 & 2000 &  0.054 & 0.066 && 0.058 & 0.070 &0.104 & 0.102 & 0.528 && 1.000& 1.000 && 1.000  & 1.000 & 0.558 & 1.000 & 1.000 \\
							\\   \multirow{2}{*}{M6} & 1000 & 1000 &  0.056 & 0.036 && 0.050 & 0.058 & 0.055  &  0.208 & 0.558 &&   0.996 & 1.000 && 0.996 & 1.000 & 0.384 & 1.000 & 1.000    \\
							& 2000 & 2000 &  0.048 & 0.064 && 0.052 & 0.066 & 0.046 & 0.178 & 0.452 &&  1.000 & 1.000  && 1.000 & 1.000 & 0.362 & 1.000 & 1.000   \\
							\hline
						\end{tabular}
						\footnotesize Notes: GCA-CDET(NN) and GCA-CDT (LLR) is the proposed method with the classifier learned  by neural networks and linear logistic regression, respectively.  GCIT is the Generative Conditional Independence Testing method proposed by \cite{bellot2019conditional}. DRGCIT is the Doubly Robust Generative Conditional Independence Testing method proposed by \cite{zhang2025doubly}.
						WCPT is the weighted conformal prediction methods by \cite{hu2023two}.
				\end{minipage}}
		\end{threeparttable}}
	\end{table}

	\begin{table}[h]{\color{black}
			\caption{Average running time per replication (in seconds) when $n_1 = n_2 = 1000$.}\label{tab:time}
			\begin{threeparttable}
				\resizebox{\textwidth}{!}{\begin{minipage}{\textwidth}
						\centering
						\begin{tabular}{c cc cc cc}
							\hline
							& M1 & M2 & M3 & M4 & M5 & M6  \\
							\hline
							GCA-CDET & 44.65 (1.85) & 48.82 (1.76)  & 45.76 (2.09)  & 52.24 (1.82)  & 66.78 (1.63)  & 62.92 (2.47)  \\
							GCIT & 55.94 (1.57)  & 70.65 (1.98) & 73.18 (0.99) & 85.83 (1.32) & 92.39 (2.88) & 88.35 (2.96) \\
							DRGCIT & 416.96 (3.39) & 441.61 (3.98) & 448.91 (4.42) & 462.44 (3.66) &  453.58 (3.69) & 1832.47 (3.70)\\
							WCPT &  85.04 (1.43) & 84.94 (1.49) & 96.58 (1.65) &  96.81 (1.33) & 253.42(1.85) & 105.34(1.27)\\
							\hline
						\end{tabular}
				\end{minipage}}
				\footnotesize Notes: All experiments (except WCPT) were conducted in Python using an NVIDIA RTX 4090 GPU. The WCPT experiments were implemented in R and executed on an Xeon 6240 CPU.
		\end{threeparttable}}
	\end{table}
	
	\subsection{Real data-based simulation}\label{syn_sim}
	To investigate the performance of the proposed method in handling complex covariate distributions in real world application, we design a simulation based on a real dataset: the Bike Sharing dataset \citep{misc_bike_sharing_275}, which contains 17,379 observations, with 12 covariates comprising both discrete and continuous variables. The response $Y$ is the hourly count of bike rentals. Data standardization is applied.  As the computation time of DRGCIT is considerably longer for such a large dataset,  in this experiment, we only compare GCA-CDET with GCIT and WCPT.
	The implementation details are presented in the supplementary material.
	
	As the original data do not form a two-sample problem, we construct synthetic datasets by randomly selecting	 $n_1$ and $n_2$ samples from the dataset, denoted as $\mathbb{D}_1$ and $\mathbb{D}_2$, respectively. Different sample sizes $n_1$ and $n_2$ are tried as shown in Table \ref{tab:bike}. 
	We consider two experiments:
	(i) directly apply the testing methods to $\mathbb{D}_1$ and $\mathbb{D}_2$; 
	(ii) apply the testing methods to $\mathbb{D}_1$ and a modified $\mathbb{D}_2$
	by adding each observation of $Y$  in  $\mathbb{D}_2$  by 0.5. 
	In experiment (i), 	 
	due to the nature of random sampling,  the null hypothesis $\mathbb{P}_{1,Y|X}=\mathbb{P}_{2,Y|X}$ holds; 	 
	in experiment (ii), the alternative hypothesis $\mathbb{P}_{1,Y|X}\neq\mathbb{P}_{2,Y|X}$ is true due to the artificial modification on  $Y$ in $\mathbb{D}_2$.
	
	The results in  Table \ref{tab:bike} are based on 100 repetitions with a nominal significance level  of $\alpha=0.05$. 
	Table \ref{tab:bike} shows that, in experiment (i),  type I errors are well controlled for all methods.
	Moreover, the proposed GCA-CDET is  robust in controlling the type I error  for the imbalanced case with empirical type I errors closer to the nominal level, especially for  scenarios with large $n_1$ and small $n_2$ such as $n_2\in\{1000,2000\}$.
	In experiment (ii), all methods except GCIT achieve a power of 1.00 across all settings,  indicating  strong evidence against the null hypothesis.

	\begin{table}[h]{\color{black}
			\centering
			\caption{Percentage of rejections over 100 repetitions on the Bike Sharing dataset.}
			\label{tab:bike}
			\begin{threeparttable}
				\resizebox{\textwidth}{!}{\begin{minipage}{\textwidth}
						\begin{tabular}{cc c cc c ccccc c c }
							\hline
							& &\multicolumn{4}{c}{Experiment (i)} && \multicolumn{4}{c}{Experiment (ii)}\\ 
							\cline{3-6}\cline{8-11}
							&  & \multicolumn{2}{c}{GCA-CDET} &   \multirow{2}{*}{GCIT}  &   \multirow{2}{*}{WCPT} && \multicolumn{2}{c}{GCA-CDET} &   \multirow{2}{*}{GCIT}   &  \multirow{2}{*}{WCPT} \\
							\cline{3-4} \cline{8-9}
							$n_1$ & $n_2$ & NN & LLR &&&  & NN& LLR &&  \\
							\hline
							8689 & 1000 & 0.05 & 0.06 & 0.09 &  0.08 &&1.00&1.00& 0.26 &1.00 \\
							& 2000 & 0.04 & 0.07 & 0.09 & 0.11 && 1.00&1.00& 0.31 &1.00 \\
							& 5000 & 0.07 & 0.04 & 0.08 & 0.05 && 1.00&1.00& 0.28 &1.00 \\
							& 8689 & 0.08 & 0.07 & 0.08& 0.06  & &1.00&1.00& 0.44 &1.00\\
							\\
							10000 & 1000 & 0.06 & 0.04 & 0.09 & 0.10 && 1.00&1.00& 0.34 &1.00\\
							& 2000 & 0.05 & 0.01  & 0.08 &  0.04 & &1.00&1.00&0.38 &1.00 \\
							& 5000 & 0.07 & 0.06 & 0.10 &  0.05 & &1.00&1.00& 0.52&1.00 \\
							& 7379 & 0.07 & 0.02 &0.08 &  0.04 & &1.00&1.00& 0.45 &1.00\\
							\\
							15000 & 1000 & 0.05 & 0.08 & 0.12 & 0.03 && 1.00&1.00&0.40 &1.00\\
							& 2000 & 0.05 & 0.06 & 0.12 & 0.08 & &1.00&1.00&0.49 &1.00 \\
							& 2379 & 0.05 & 0.06 & 0.10 & 0.03 & &1.00&1.00& 0.51  &1.00\\
							\hline
						\end{tabular}				
				\end{minipage}}
		\end{threeparttable}}
	\end{table}

	\section{Real Data Analysis}\label{sec:data}
	Two real datasets are considered in this section: Wine Quality dataset \citep{Cortez2009ModelingWP} and HIV-1 Drug Resistance dataset \citep{rhee2006genotypic}.  
	The implementation details are provided in Section S.9 of the online supplementary materials.
	All tests are conducted at a significance level $\alpha=0.05$.
	\subsection{Wine Quality dataset} 
	
	The Wine Quality dataset consists of 6497 samples, among which  4,898 samples are white wine and 1,599 samples are red wine.
	Eleven physicochemical variables are collected  and treated as covariates.
	The response variable is a sensory score, which measures the wine quality and ranges between 0 and 10. 
	
	To examine whether the relationship between physicochemical variables and sensory score is the same across two wine types, we apply the testing methods and compute the corresponding $p$-values.
	The $p$-values of all methods are below $0.05$, suggesting rejection of the null hypothesis. This implies that the relationship between physicochemical variables and sensory scores  is different  between the two wine types.
	Especially, the proposed GCA-CDET with NN and LLR yields $p$-values of  $5.48\times10^{-33}$ and $2.51\times 10^{-57}$, which are comparable to that of DRGCIT ($0.000$), and substantially smaller than those  by GCIT ($0.006$) and WCPT ($0.020$). 

	To provide a more comprehensive analysis, we conduct two additional experiments.
	In experiment (i), the white wine dataset is randomly partitioned into two subsets of sizes $n_1$ and $n_2$, respectively. As the relationship between physicochemical variables and sensory scores remains the same across the two subsets, the null hypothesis holds.
	In experiment (ii), $n_1$ and $n_2$ samples are randomly selected from the white and red wine datasets, respectively. The previously obtained $p$-values suggest that the alternative hypothesis is true.
	Different sizes $n_1$ and $n_2$ are considered in both experiments, and 100 trials are conducted. Results are summarized in Table \ref{tab:wine}, 
	which shows that,  the proposed methods achieve valid type I error control across different sample sizes, with empirical type I errors close to the nominal level, whereas the other methods do not.
	Moreover, in experiment (ii), GCA-CDET generally gives larger empirical powers against GCIT and WCPT, showing the advantage of the GCA-CDET in power comparison.

	\begin{table}[h]{\color{black}
			\centering
			\caption{Percentage of rejections over 100 trials on the Wine Quality dataset.}
			\label{tab:wine}
			\begin{threeparttable}
				\resizebox{\textwidth}{!}{\begin{minipage}{\textwidth}
						\begin{tabular}{cc c cc c cc c c cc cc c cc c cc }
							\hline
							&&\multicolumn{5}{c}{Experiment (i)} &&  &&\multicolumn{5}{c}{Experiment (ii)}\\ 
							\cline{3-7} \cline{11-15}
							&  & \multicolumn{2}{c}{GCA-CDET} & \multirow{2}{*}{GCIT} & \multirow{2}{*}{DRGCIT} & \multirow{2}{*}{WCPT} && & & \multicolumn{2}{c}{Proposed} & \multirow{2}{*}{GCIT}& \multirow{2}{*}{DRGCIT} & \multirow{2}{*}{WCPT} \\
							\cline{3-4} \cline{11-12}  
							$n_1$ & $n_2$ & NN & LLR &&  &   && $n_1$ & $n_2$ & NN& LLR && &  \\
							\hline
							2000 
							& 1000 & 0.05 & 0.01 & 0.13& 0.32 & 0.11 &&2000 & 1000 & 0.95 & 0.88& 0.54 & 1.00 & 0.18\\
							& 2000 & 0.07 & 0.02 & 0.18 & 0.28 & 0.11 && & 1500 & 0.98 & 0.92& 0.50 & 1.00 & 0.38\\
							\\ 2500 
							& 1000 & 0.05&0.03 & 0.14& 0.25 & 0.11 && 3000& 1000 & 0.96 & 0.94& 0.55& 1.00 & 0.18 \\
							& 2000 & 0.04&0.03 &0.14& 0.27 & 0.09 && & 1500 & 0.99 & 0.95& 0.56 & 1.00 & 0.38 \\
							\\ 3000 
							& 1000 & 0.04 & 0.02 &0.16& 0.26  & 0.08 & & 4000 & 1000 & 0.97 & 0.97& 0.57 & 1.00 & 0.18 \\
							& 1898 & 0.07 & 0.03 &0.12& 0.22 & 0.14 && & 1500 & 1.00 & 0.96&0.64 & 1.00 & 0.38\\
							\hline
						\end{tabular}
				\end{minipage}}	
		\end{threeparttable}}
	\end{table}

	\subsection{HIV-1 Drug Resistance Dataset}
	
	This dataset contains information on 16 drugs from three classes.
	We  consider two classes: the nucleotide reverse transcriptase inhibitors (NRTIs) to which nine drugs belong, and the non-nucleoside RT inhibitors (NNRTIs)  containing three drugs. In this analysis, the  response $Y$ is the log-transformed drug resistance level.
	Each component of $X$ is a binary indicator of mutation presence.
	After removing samples with missing drug resistance data and rare mutations (fewer than three occurrences), the final dataset contains 319 covariates and 5,718 observations.
	The sample sizes for the nine drugs are as follows: 633 for 3TC, 628 for ABC, 630 for AZT, 632 for D4T, 353 for DDI, 732 for DLV, 734 for EFV, and 746 for NVP.
	
	The primary goal of this  analysis is to examine whether the distribution of drug resistance levels given the gene mutations are identical across the two drug classes NRTIs and NNRTIs, which is the null hypothesis in this analysis.
	For different methods,  their $p$-value are computed and presented in Table \ref{tab:HIV}, which  
	shows that almost all $p$-values of the GCA-CDET,GCIT and DRGCIT are much smaller than 0.05, suggesting rejection of the null hypothesis.
	This decision is consistent with the finding of a  medical study  \citep{rhee2006genotypic}  that,  drugs within these two classes target different gene mutations through distinct mechanisms.
	Conversely, WCPT  gives  $p$-values much larger than 0.05 in some cases, leading to 
	an opposite decision.

	\begin{table}[h]{\color{black}
			\caption{P-value for HIV-1 dataset.} \label{tab:HIV}
			\begin{threeparttable}
				\resizebox{\textwidth}{!}{\begin{minipage}{\textwidth}
						\begin{tabular}{c c ccc ccc}
							\hline
							&  & \multicolumn{6}{c}{NRTIs}\\
							\cline{3-8}
							Methods & NNRTIs & 3TC & ABC & AZT & D4T & DDI & TDF \\
							\hline
							GCA-CDET (NN) & DLV & $3.54\times10^{-19}$ & $6.09\times 10^{-21}$ & $4.20\times10^{-4}$ & $5.75\times10^{-8}$ & $8.92\times10^{-7}$ & $0.003$ \\
							& EFV & $4.12\times10^{-19}$ & $9.27\times10^{-34}$ & $2.66\times10^{-8}$  & $4.52\times10^{-14}$ & $3.93\times10^{-17}$ & $4.00\times10^{-5}$ \\
							& NVP & $1.14\times10^{-11}$ & $2.13\times10^{-38}$ & $4.13\times10^{-7}$ & $2.90\times10^{-10}$ & $3.07\times10^{-12}$ & $3.06\times10^{-4}$ \\
							\\
							GCA-CDET (LLR) & DLV & $5.35\times10^{-11}$ & $2.77\times10^{-8}$ & 0.035 & $4.75\times10^{-7}$ & $4.74\times10^{-7}$ & $5.22\times10^{-4}$ \\
							& EFV & $4.75\times10^{-12}$ & $1.50\times10^{-6}$ & 0.027 & $1.62\times10^{-6}$ & $9.98\times10^{-4}$ & $4.86\times10^{-4}$  \\
							& NVP &  $2.52\times10^{-9}$ & $2.51\times10^{-10}$ & 0.006 & $1.56\times10^{-11}$ & $2.32\times10^{-13}$ & $2.34\times10^{-4}$   \\
							\\
							GCIT & DLV & 0.001 &  0.022 & 0.001 & 0.007 & 0.020 &  0.000\\
							& EFV & 0.002 & 0.001 & 0.012 & 0.008 & 0.045 & 0.044\\
							& NVP & 0.009 & 0.262 & 0.146 & 0.009 & 0.018 & 0.017 \\
							\\
							DRGCIT & DLV& 0.000 & 0.000 & 0.000 & 0.000 & 0.000 & 0.000 \\
							&EFV & 0.000 & 0.000 & 0.000 & 0.000 & 0.000 & 0.000\\
							& NVP & 0.000 & 0.000 & 0.000 & 0.000 & 0.000 & 0.000\\
							\\
							WCPT & DLV & 0.005 & 0.072 & 0.390 & 0.032 & 0.019 & 0.067 \\
							& EFV & 0.002 & 0.067 & 0.181 & 0.046 & 0.031 & 0.060\\
							& NVP & 0.011 & 0.017 & 0.342 & 0.008 & 0.005 & 0.028\\
							\hline
						\end{tabular}
				\end{minipage}}
		\end{threeparttable}}
	\end{table}

	\section{Discussion}\label{Discussion}
	In this paper, we propose a general and flexible framework for testing the equality of two conditional distributions based on conditional generative learning methods using deep neural networks. To illustrate the proposed framework, we develop the generative classification accuracy-based conditional distribution equality test using MDNs. We investigate the convergence properties of the learned MDN conditional generator and prove the testing consistency for GCA-CDET.  To validate the proposed methods empirically,  we compute  the  GCA-CDET within our proposed framework through various numerical experiments. Furthermore, in the online supplementary material, we establish a minimax lower bound for statistical inference of testing the equality of two conditional distributions under certain smoothness conditions, and demonstrate that some tests within the proposed framework achieve this lower bound, either exactly or up to an iterated logarithm factor.
	
	Several questions deserve further investigation. For example:  (1) Randomness of data splitting and synthetic data generation. The proposed tests are intrinsically randomized due to the data splitting for $\mathbb{D}_2$ and the data generation through the learned conditional generator and $\mathbb{D}_1$. Without carefully documenting random seeds, researcher can “select” the results by reporting the best results across different splits and synthetic data generation. 
	A possible solution to this issue is to apply de-randomized techniques  based on e-values, which have been used in a series of recent work such as \cite{ren2023derandomizing,bashari2024derandomized,ren2024derandomised}. These authors have shown that using e-values successfully stabilizes the output of some randomized tests. (2) Considering other  conditional generative learning approaches.  In this work, we use MDNs for learning the conditional generator. It would be  interesting  to consider other state-of-the-art conditional generative learning approaches, such as  conditional stochastic interpolation,  conditional F\"ollmer flow, and conditional diffusion model. We leave these questions for future work.

	\section*{Acknowledgment}
	All the data and code for the numerical experiments will be public. 
	Two real datasets can be downloaded from: \url{https://archive.ics.uci.edu/dataset/186/wine+quality} for  Wine Quality dataset, and \url{https://hivdb.stanford.edu/pages/published_analysis/genophenoPNAS2006/} for HIV-Drug Resistance dataset.

\begin{singlespace}
\bibliographystyle{apalike}
\bibliography{Reference}
\end{singlespace}
\end{document}